\documentclass[10pt,twocolumn,letterpaper]{article}

\usepackage[pagenumbers]{cvpr} 

\usepackage{graphicx}
\usepackage{amsmath}
\usepackage{amssymb}
\usepackage{booktabs}

\usepackage[pagebackref,breaklinks,colorlinks]{hyperref}

\usepackage[capitalize]{cleveref}
\crefname{section}{Sec.}{Secs.}
\Crefname{section}{Section}{Sections}
\Crefname{table}{Table}{Tables}
\crefname{table}{Tab.}{Tabs.}

\def\cvprPaperID{10} 
\def\confName{CVPR}
\def\confYear{2023}

\begin{document}

\title{Investigating Catastrophic Overfitting in Fast Adversarial Training: A Self-fitting Perspective}

\author{Zhengbao He \quad Tao Li \quad Sizhe Chen \quad Xiaolin Huang\thanks{This work was partially supported by National Natural Science Foundation of China (61977046), Research Program of Shanghai Municipal Science and Technology Committee (22511105600), and Shanghai Municipal Science and Technology Major Project (2021SHZDZX0102). Xiaolin Huang is the corresponding author.}\\
Department of Automation, Shanghai Jiao Tong University\\
{\tt\small \{lstefanie, li.tao, sizhe.chen, xiaolinhuang\}@sjtu.edu.cn}
}
\maketitle

\begin{abstract}
Although fast adversarial training provides an efficient approach for building robust networks, it may suffer from a serious problem known as catastrophic overfitting (CO), where multi-step robust accuracy  suddenly collapses to zero. In this paper, we for the first time decouple single-step adversarial examples into data-information and self-information, 
which reveals an interesting phenomenon called ``\emph{self-fitting}''. Self-fitting, i.e., the network learns the self-information embedded in single-step perturbations, naturally leads to the
occurrence of CO.
When self-fitting occurs,  the network experiences an obvious ``\emph{channel differentiation}'' phenomenon that some convolution channels accounting for recognizing self-information become dominant, while others for data-information are suppressed. In this way, the network can only recognize images with sufficient self-information and loses generalization ability to other types of data. 
Based on self-fitting, we provide new insights into the existing methods to mitigate CO and extend CO to multi-step adversarial training.
Our findings reveal a self-learning mechanism in adversarial training and open up new perspectives for suppressing different kinds of information to mitigate CO.
\end{abstract}

\section{Introduction}
\label{sec:intro}

Deep neural networks (DNNs) suffer from a significant threat from adversarial attacks, which deceive DNNs by adding invisible perturbations to the inputs \cite{szegedy2013intriguing}. 
In this regard, how to defend DNNs against such malicious attacks has attracted much attention \cite{xie2017mitigating, gong2017adversarial, madry2017towards, chen2020universal, chen2022relevance}. Adversarial training (AT), which directly augments the datasets with adversarial examples (AEs), is considered one of the most effective defense methods \cite{madry2017towards}. 
However, standard AT consumes great computation as it involves multiple forward and backward propagations to generate adversarial examples.

To address such prohibitive burdens, single-step adversarial training (also known as fast adversarial training, FAT \cite{wong2020fast}) is proposed to generate adversarial examples efficiently via simply one-step gradient propagation. 
Unfortunately, a severe issue  arises  that during the training where multi-step (PGD) robust accuracy \cite{madry2017towards} can suddenly drop to zero within a few epochs, while the single-step robust accuracy increases rapidly. Such dramatic phenomenon is referred as \emph{“catastrophic overfitting”} (CO) \cite{wong2020fast}.

\begin{figure}[!t]
\centering
\includegraphics[width=0.48\textwidth]{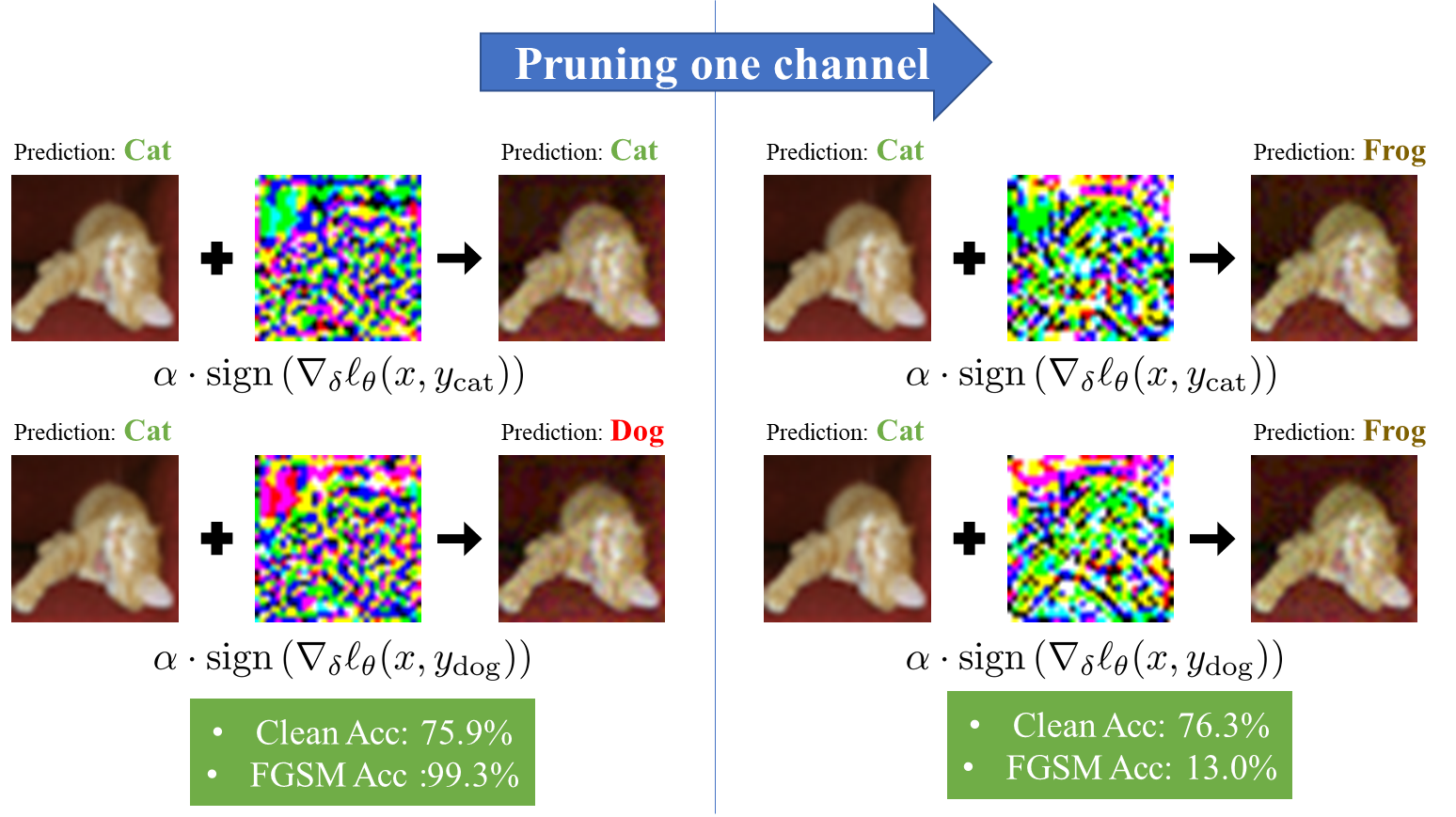}
\caption{
Illustration
of self-fitting and channel differentiation phenomenon.
\textbf{(Left)}
A network with CO has a clean accuracy of 75.9\% and an FGSM accuracy of 99.3\%. The prediction of FGSM examples is determined by the label used to generate single-step perturbation, which suggests that the DNN learns to recognize the self-information that the network embeds in the single-step perturbation. \textbf{(Right)}
After pruning only one channel, the perturbation becomes adversarial again, leading to a drop of 86.3\% in FGSM accuracy, while the clean accuracy only increases by 0.4\%. The pruned channel recognizes FGSM examples which contain self-information while others recognize clean examples, suggesting that different parts fit different kinds of information. }
\label{fig:sketch}
\end{figure}
Currently, methods for resolving CO can be broadly  classified into two categories: generating diverse AEs \cite{zheng2020efficient, kim2021understanding,chen2022efficient} and applying proper regularization techniques \cite{andriushchenko2020understanding, sriramanan2020guided}. 
These methods are designed based on a consensus that DNNs lose robustness against multi-step attacks due to overfitting to single-step adversarial examples.
More specifically, most of them have either directly or implicitly argued that the generated adversarial perturbations are ``meaningless'' after CO,  rendering them to be ineffective for model robustness \cite{andriushchenko2020understanding, ortiz2022catastrophic}.
There are also some recent works studying CO from a data-feature perspective, trying to create a connection between single-step AEs and the intrinsic properties of data distribution \cite{yang2023revisiting}.
However, the effects of model information on generating AEs have not been studied, and we will demonstrate its importance in our research.

In this paper, we decouple single-step AEs after CO into two parts: the data-information part and the self-information part. 
Based on this, we then interpret CO from a new \emph{“self-fitting”} perspective that DNN eventually learns to recognize the self-information embedded in the single-step adversarial perturbations.
Instead of being harmful, such self-information conversely helps the network recognize single-step AEs, which were considered meaningless previously.
For example, on a ResNet18 \cite{he2016deep} trained on Cifar10 \cite{krizhevsky2009learning}, even if the target labels are completely shuffled, the network still holds an FGSM (Fast Gradient Sign Method \cite{43405}) robust accuracy of 90\% after CO, which should be under 10\% for a normal trained network due to  meaningless labels and attack operations.
This suggests that after CO, the network makes predictions mainly based on adversarial perturbations, i.e., the self-information, instead of the data information. In this way, the network prediction is consistent with the random labels for generating AEs.

In more detail, DNNs use different structures to learn different information 
\cite{patil2021convolutional, ismail2019deep, NIPS2017_3f5ee243}.
We find that this is also true for self-fitting such that there are special channels for fitting self-information, namely \emph{"channel differentiation"}.
Specifically, channels with high variance  mainly recognize attack information, i.e., the self-information,
while other channels identify clean examples. As shown in \cref{fig:sketch}, by pruning only one channel with the highest variance, the FGSM accuracy of a ResNet18 with CO decreases from 99.3\% to 13.0\%, while the clean accuracy increases from 75.9\% to 76.3\%. That is this channel is mainly in charge of recognizing the self-fitting information.

From the view of self-fitting, we further give new insights into the existing methods to mitigate CO. For example, only adversarial initialization cannot prevent CO while proper regularization is necessary. 
We could  postpone the happening of CO by suppressing channels that recognize self-fitting information. 
We also extend CO to multi-step AT, revealing that CO can also happen in multi-step AT with few iterations or large step sizes.
Besides, the channel differentiation provides a new starting
point for studying single-step adversarial training from the perspective of network parameters, which can help us explore the training dynamics.
Our contributions can be summarized as follows:
\begin{itemize}
    \item We propose \emph{"self-fitting"}, a novel perspective to interpret catastrophic overfitting in fast adversarial training.
    \item We delve into the changes of network structure during CO, pointing out that there are some special channels for fitting self-information, i.e., the \emph{"channel differentiation"} phenomenon. 
    \item Based on self-fitting, we provide new insights into previous methods for mitigating CO, and also extend CO to multi-step AT. 
\end{itemize}

\section{Related work}

\subsection{Adversarial Attack}

Since DNNs are discovered vulnerable to adversarial examples  \cite{szegedy2013intriguing}, a family of methods are proposed to generate AEs. FGSM \cite{43405} and PGD \cite{madry2017towards} are two popular white-box attack algorithms that craft AEs according to the gradient of loss w.r.t. to input. Square attack \cite{andriushchenko2020square} is a state-of-the-art black-box attack which iteratively crafts adversarial examples without gradients by only querying the DNN's outputs, and there have been special defense for query attacks \cite{chen2022adversarial, wu2022unifying}. AutoAttack (AA)  ensembles four attack algorithms, including two white-box attack algorithms and two black-box attack algorithms  \cite{croce2020reliable}. AA is recognized as one of the strongest attack algorithms and is widely used to evaluate adversarial robustness.

\subsection{Efficient Adversarial Training}

Adversarial training  \cite{madry2017towards, li2022subspace} is considered to be the most effective way to defend against adversarial attacks by augmenting training data with adversarial examples. Since the generation of AEs is time-consuming, many variants of AT try to improve training efficiency. 
Free AT \cite{shafahi2019adversarial} updates network parameters while generating AEs at the same time. 
YOPO  \cite{zhang2019you} restricts most of the forward and back propagation within the first layer of the network during AE updates. 
Fast AT  \cite{wong2020fast} replace multi-step attacks with FGSM and generate AEs with single-step gradient.

\subsection{Catastrophic Overfitting}

Although Fast AT can achieve comparable performance to multi-step AT, it suffers from catastrophic overfitting, i.e., the PGD accuracy drops to zero while the FGSM accuracy increases quickly within a few epochs.
Some works mitigate CO by generating more adversarial examples. For example, ATTA \cite{zheng2020efficient} combines AEs of the last epoch and gradient of the current epoch, and N-FGSM \cite{de2021towards} generates AEs using a stronger noise without clipping the perturbation. 
Other works apply strong regularization terms to stabilize the training. For example, GradAlign \cite{andriushchenko2020understanding} prevents CO by maximizing the gradient alignment around clean examples to reserve local linearity. 
Compared to the consensus that the adversarial perturbations become "meaningless" after CO, we reveal that they contain much model information (self-information) to conversely facilitate the network recognition.

\section{Self-fitting phenomenon in FAT}
Standard adversarial training  \cite{madry2017towards} can be formatted as a min-max problem as:
\begin{equation}
  \min_{\theta}  ~  \mathbb E_{( x, y) \in \mathcal D} \left [ \max_{\delta \leq \epsilon} \ell_{\theta}(x+\delta, y) \right  ],
  \label{eq:standard-at}
\end{equation}
where $\theta \in \mathbb R^n$ denotes network parameters; $y$ is the ground-truth label of sample $x  \in  \mathbb R^d$ from training set $\mathcal D$; $\delta \in  \mathbb R^d$ is the adversarial perturbation constrained with the norm bound $\epsilon$; $x + \delta$ denotes the adversarial sample; $\ell_{\theta}(\cdot)$ is the loss function correspond to the specific parameter $\theta$.

In FAT, for a certain sample pair $(x, y)$, the optimized objective can be written as:
\begin{equation}
  \min_{\theta}  ~ \ell_{\theta} (x + \alpha  \cdot \delta(x, \theta), y),
  \label{eq:fat}
\end{equation}
where $\delta(x, \theta) =  \mathrm{sign} \left ( \nabla_{x} \ell_{\theta}(x, y)\right )$. Origin data $x$ control data information while model parameters $\theta$ control information from the network, i.e., self-information. Note that different $y$ correspond to different weights in the linear layer of $\theta$. As a result, $y$ can also control the self-information based on $\theta$.

Normal adversarial training should balance the information from both data and the network itself. The former helps the network to learn meaningful knowledge from the dataset while the latter enables the network to resist a wide range of unknown attacks. 
If $\delta$ is only determined by $\theta$, the network may only reserve robustness to specific attacks, such as universal adversarial training  \cite{9428419} where different $x$ share a common $\delta$. If $\delta$ is only determined by $x$ and independent with $\delta$, the network can only gain black-box robustness, as discussed by  \cite{yang2020dverge}. Thus, fitting the information of the network with proper $\delta$ is the key for AT to achieve white-box robustness. 

However, it is hard to decouple the two kinds of information since $x +\alpha \cdot \delta(x,y) $ is sent back to network as a whole when performing the model parameter optimization. 
When $\alpha$ is large enough, $\delta$ may become the main basis for network classification.
If the network pays much attention to its self-information in $\delta$, the network may overfit one specific attack. As a result, the prediction is determined by only self-information and is irrelevant to the data-information. 
In other words, the perturbation is not harmful but useful for the network to recognize adversarial examples. This shortcut solution has a low loss value but is hard to generalize to other attacks or even the original dataset because data-information is unimportant for classification under such a shortcut solution. We name the possible shortcut solution as "self-fitting".

The above analysis illustrates the possible existence of self-fitting in adversarial training, which may lead to abnormal behavior during the training process.
In this section, we probe this experimentally and show that there is a connection between self-fitting and catastrophic overfitting (CO).
We argue that self-fitting could be a primary underlying cause for the occurrence of CO --- $\delta$ contains the label information $y$ and eventually dominates the predictions.

\begin{figure}[!t]
\centering
\includegraphics[width=0.4\textwidth]{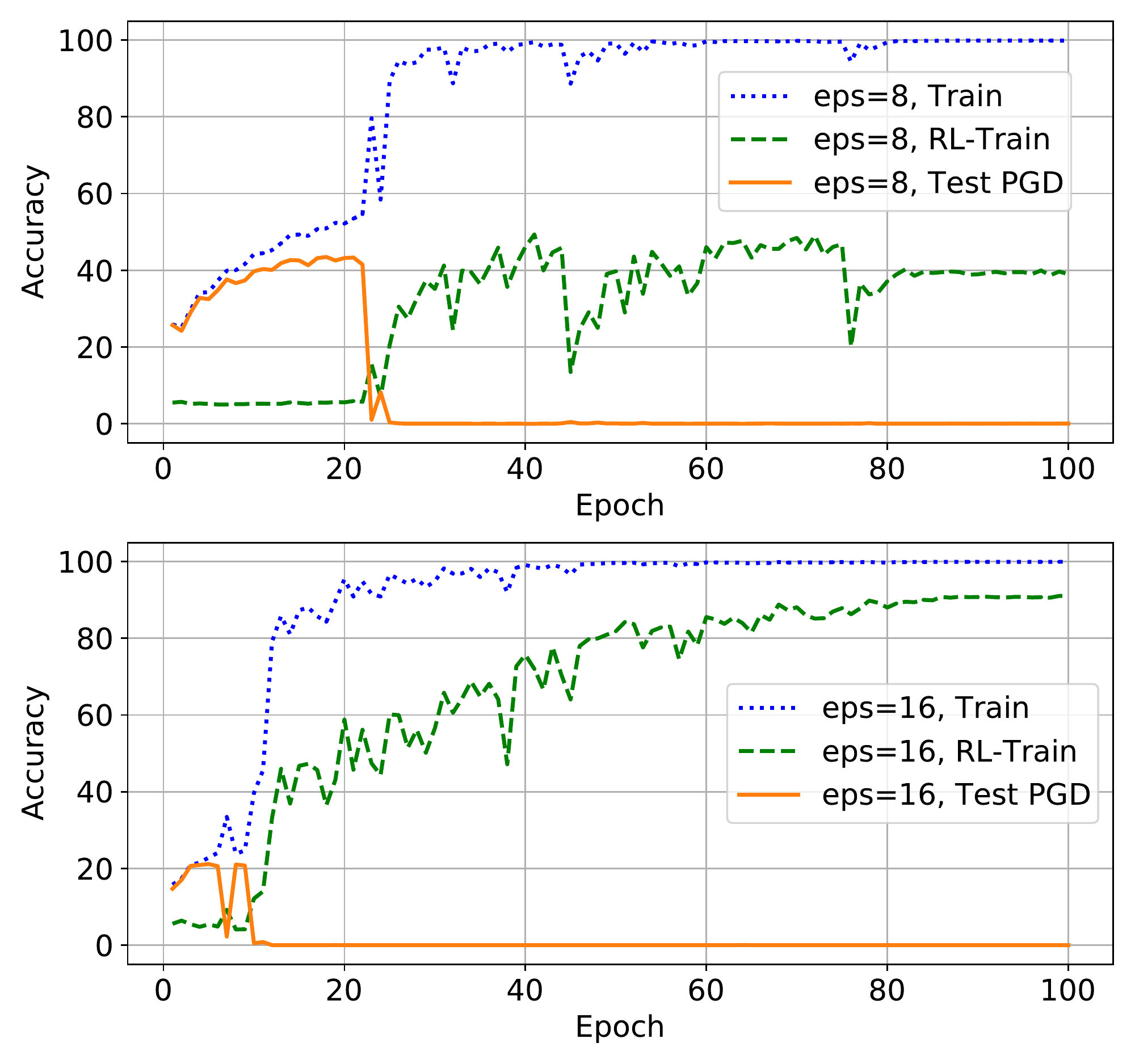}
\caption{FGSM-AT training with different $\epsilon$ on Cifar10 using ResNet18. Catastrophic overfitting happens at 23rd epoch for $\epsilon = 8/255$ and 10th epoch for $\epsilon = 16/255$. Line "Train" accuracy is the FGSM accuracy of the training set. Line "RL-Train" accuracy is the random-label FGSM accuracy of the training set. Line "Test PGD" accuracy is the PGD10 accuracy of the test set.}
\label{fig:base-training}
\end{figure}

In order to demonstrate the self-fitting phenomenon in CO, we first 
 train ResNet18 on Cifar10 with FGSM-AT method  \cite{wong2020fast} with different $\epsilon$. The training curve is shown in \cref{fig:base-training}. Catastrophic overfitting happens at 23rd epoch for $\epsilon = 8/255$ and 10th epoch for $\epsilon = 16/255$.

 The following experiment is designed to observe if the prediction changes with the self-information change. Since $y$ can also control the self-information in $ \mathrm{sign} \left ( \nabla_{\delta} \ell_{\theta}(x, y)\right )$, we fix origin samples $x_{\text{ori}}$ and change labels $y$ to generate adversarial perturbations with  different self-information and test the random-label FGSM accuracy  (RL-FGSM accuracy) of a network after CO. Practically, we first assign each sample a random label before testing, and then generate the adversarial example w.r.t. that random label, i.e.:
\[
x_{\text{random}} = x_{\text{ori}} + \alpha \cdot \mathrm{sign} \left ( \nabla_{\delta} \ell_{\theta}(x_{\text{ori}}, y_{\text{random}})\right ),
\]
where $\alpha = \epsilon$ is the attack step size. Then accuracy is calculated by comparing the prediction $f_\theta(x_\text{random})$ and random label $y_{\text{random}}$, where $f_\theta(\cdot)$ represents the parametric model with $\theta$.
For a normal trained model, the accuracy on a completely shuffled dataset should be 10\%. After FGSM attack, the accuracy is anticipated to drop below 10\%. We test the RL-FGSM accuracy of checkpoints saved at the end of each epoch for different $\epsilon$. The result is shown in \cref{fig:base-training}. We show results using the training set, but similar ones can be observed in the test set. 

\begin{figure}[!t]
\centering
\includegraphics[width=0.4\textwidth]{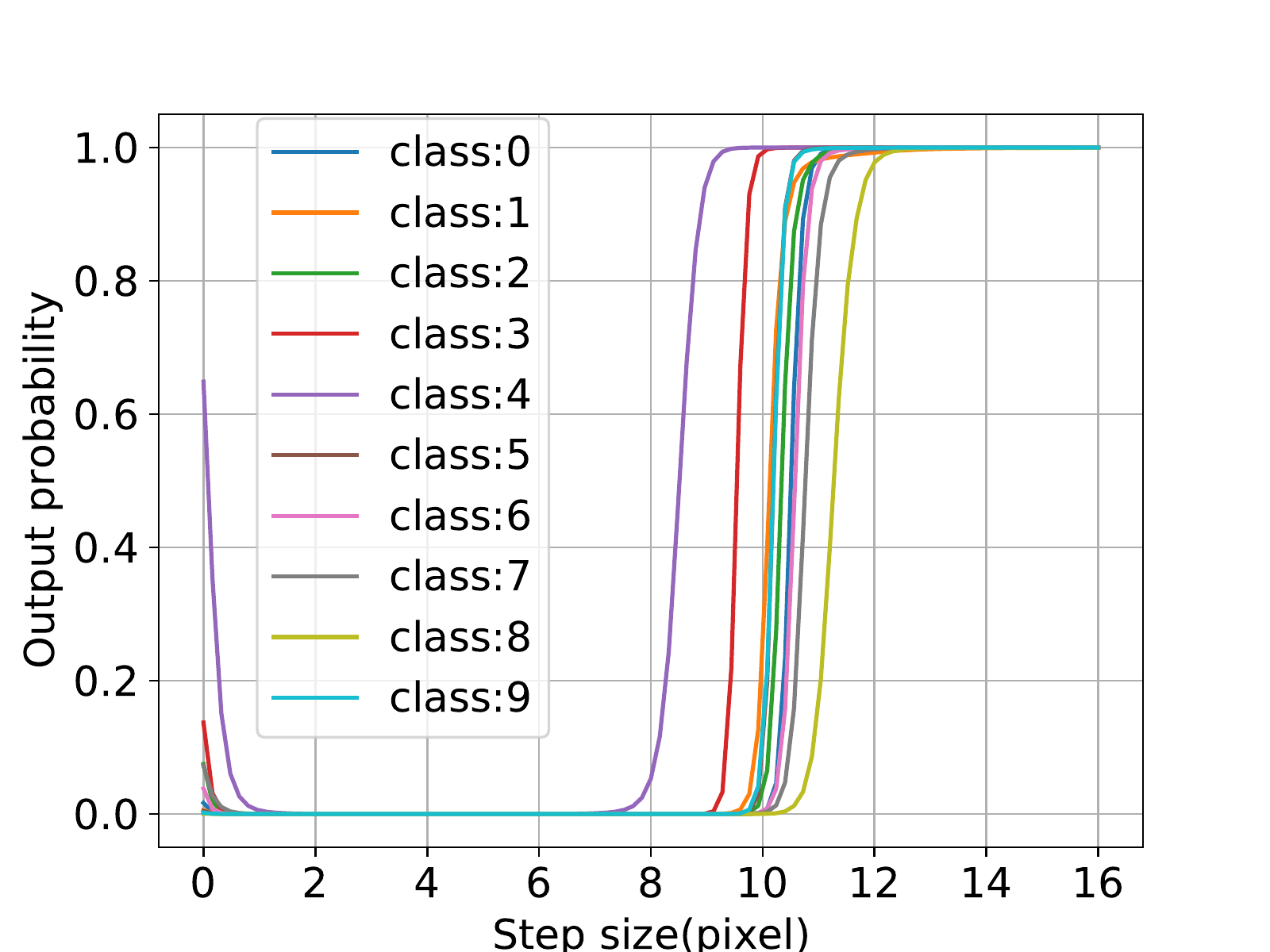}
\caption{Visualization of the probability of the network in the corresponding classes when the step size of FGSM perturbation gradually increases. The original class is class 4. The network is trained with $\epsilon = 16/255$. }
\label{fig:change-alpha}
\end{figure}

From the experiments, we can conclude that: 
\romannumeral1) The RL-FGSM accuracy of a network with CO is more than 10\%, which suggests that for those FGSM examples generated with wrong labels, the network can recognize the self-information embedded in the single-step adversarial perturbation and predict those AEs as specific random labels. We randomly choose one example from the test set and perform 10 FGSM attacks with different classes as the ground-truth. \cref{fig:change-alpha} visualizes the output probability of the network in the corresponding class when the step size of FGSM perturbation gradually increases. We observe that when the step size is small, the perturbation can fool the network to misclassify AEs, which agrees with the decision boundary distortion phenomenon discovered by  \cite{kim2021understanding}. But more surprisingly, not only the perturbation generated with the right label but also the perturbation generated with other labels can help the network classify single-step AE as the corresponding label when the step size is close to $\epsilon$.  
\romannumeral2) As catastrophic overfitting (CO) occurs, the RL-FGSM accuracy of the network gradually increases with further training, which suggests that the network alters its learning approach and starts to acquire the capability to identify self-information. Consequently, it suddenly loses robustness to other attacks.
\romannumeral3) The network trained with larger $\epsilon$ exhibits higher RL-FGSM accuracy, with a maximum value exceeding 90\%.
In contrast, the maximum value of the network trained with $\epsilon=8/255$ is only about 40\%. 
This discrepancy may be due to the fact that larger $\epsilon$  embed stronger self-information into the adversarial perturbation, making it more likely for the network to identify the self-information in single-step AEs while ignoring the information of original distribution.

Previous works  \cite{kim2021understanding, andriushchenko2020understanding, ortiz2022catastrophic} have either directly or implicitly argued that the adversarial perturbation becomes meaningless after CO so that the network  recognize single-step AEs in the same way it recognizes clean examples. However, this explanation fails to account for the low clean accuracy on original distribution after CO, especially when perturbation budget is large (for example, $\epsilon = 16/255$).  A more plausible hypothesis is that the distribution of single-step AEs after CO is distinct from the distribution of the original adversarial examples and the distribution of clean examples.
As mentioned above, an adversarial-training network may prioritize self-information in adversarial perturbation.
We think the change in adversarial distribution is the result of the network starting to learn the self-information embedded in adversarial perturbation when CO happens. 
This phenomenon that a network is capable to embed label information into adversarial perturbation  (gradient sign) and recognize the label used when generating adversarial examples could be named "self-fitting". 
Because the label information is generated through attacking the network, fitting label information of adversarial examples is fitting the network itself.

\section{Channel differentiation}

GradAlign  \cite{andriushchenko2020understanding} proposes that in a single-layer CNN, a Laplace filter that can amplify high-frequency noise grows in magnitude and outcoming weights when CO happens. Motivated by their findings, we want to further explore whether there is any part that plays a decisive role in recognizing self-information for a more complex network. 

Since the network interacts directly with the perturbation via the first layer, we concentrate our analysis on the first layer, as done in \cite{zhang2019you}. To further study the non-linear characteristics of the network, the features of all training data after the first activation layer are extracted. For example, in ResNet18, we investigate the features $f = \text{relu(bn(conv(}x_{\text{FGSM}}\text{)))}$, where $x_{\text{FGSM}}$ is the FGSM examples; $\text{conv}(\cdot)$, $\text{bn}(\cdot)$ and $\text{relu}(\cdot)$ respectively represent the first Convolution layer, the first BatchNorm layer and the first activation layer. By concatenating the features of all training examples, we can obtain a tensor with the shape of $(N,C,H,W)$, where $N$ represents the number of training examples, $C$ represents the number of channels, $H$  represents the feature height, and $W$ represents the feature width. In a ResNet18 network, $C$ is equal to 64.

\begin{figure}[!t]
\centering
\includegraphics[width=0.4\textwidth]{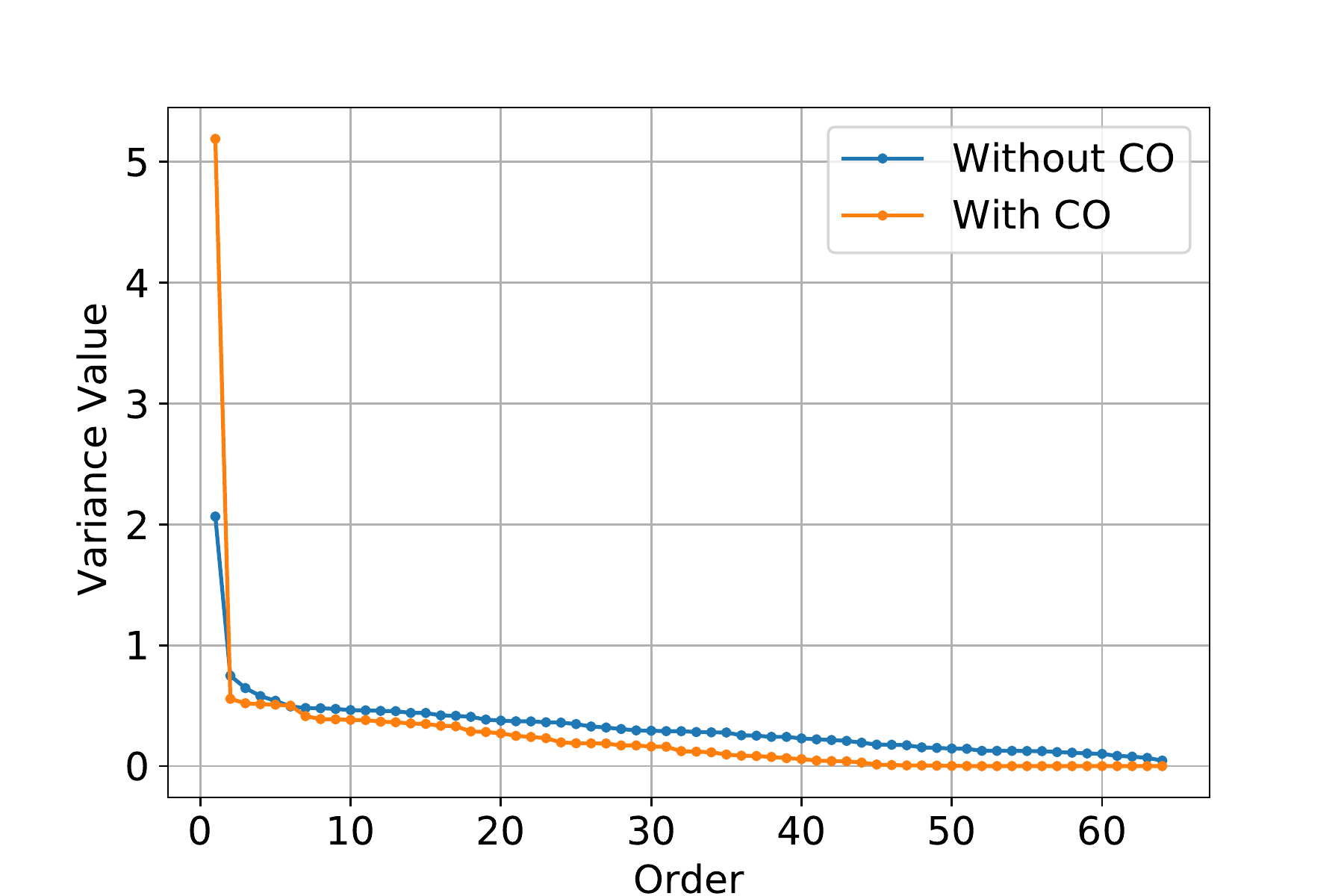}
\caption{The variance values in descending order of networks with and without the CO trained on Cifar10 using ResNet18 with $\epsilon=8/255$. The network with CO has a larger maximum variance value and more zero variance channels. }
\label{fig:channel-variance}
\end{figure}

A simple intuition is that the greater the variance of data, the more information it contains. 
Therefore, we can utilize the channel variance as a metric to evaluate the significance of different channels. Channels with higher variances are considered more crucial than those with lower variances.
We then calculate the feature variance of different channels extracted from the entire training data and sort them in descending order. The variance curves of different network checkpoints (with/without CO) in a single training experiment are illustrated in \cref{fig:channel-variance}. 
Two prominent phenomena can be observed:
\romannumeral1) The variance curve of the first few channels  becomes notably steeper after CO,  indicating a significant enhancement in the information contained in these channels. 
More specifically, the information contained in these enhanced channels could help recognize FGSM examples.
\romannumeral2) For the part of smaller variance values, there are many channels with nearly zero variance after CO. Upon closer inspection, we notice that these channels are virtually ''dead'' after CO, as their corresponding variance values approach zero. 

These two phenomena suggest that after CO, the network may become too dependent on channels with high variance after CO and might ignore other channels with lower variance.
To verify this hypothesis, we prune one channel with the highest variance and re-evaluate the clean accuracy and FGSM accuracy to investigate the role of high-variance channels in recognizing clean and FGSM examples.
This result in \cref{tabel:accuracy-change} suggests that the channels with high variance after CO are indeed crucial for the network to recognize FGSM examples. The drastic decrease in FGSM accuracy indicates that the network after CO heavily relies on these channels to recognize  FGSM examples with self-information. 
Meanwhile, the small change in clean accuracy and PGD accuracy suggests that these channels do not play a significant role in recognizing clean examples or in defending against more sophisticated attacks like PGD.

\begin{table}[ht]
\caption{Accuracy drop of different networks on Cifar10 using ResNet18 with $\epsilon=8/255$. Compared to the network without CO, the network with CO has a large drop in FGSM accuracy while little change in clean accuracy.}
\label{tabel:accuracy-change}
\vskip 0.15in
\begin{center}
\begin{small}
\begin{sc}
\begin{tabular}{crrr}
\toprule
                      & Clean & FGSM & PGD  \\ 
\midrule
w/o CO, not pruned & 75.2\%   & 50.0\% & 41.5\%   \\
w/o CO, pruned     & -0.4\%    & -2.6\%  & 0.4\%  \\ 
\midrule
 with CO, not pruned    & 75.9\%    & 99.3\%   & 0.1\%  \\
 with CO, pruned        & -0.4\%    & -86.3\%   & -2.2\% \\
\bottomrule
\end{tabular}
\end{sc}
\end{small}
\end{center}
\vskip -0.1in
\end{table}

This suggests there is a "channel differentiation" phenomenon after CO, i.e., different channels identify different kinds of examples.  Channels with high feature variance mainly recognize single-step AEs (containing much self-information) and other channels recognize clean examples. We also calculated the variance of the corresponding parameters for different channels (a large parameter variance means that this channel mainly extracts high-frequency information) and find that channels with high feature variance also have high parameter variance. In general, channels that identify high-frequency information are gradually taking a dominant role in the recognition of single-step AEs. Eventually, the over-reliance on these channels leads to the occurrence of CO.

\begin{figure}[ht]
\centering
\includegraphics[width=0.4\textwidth]{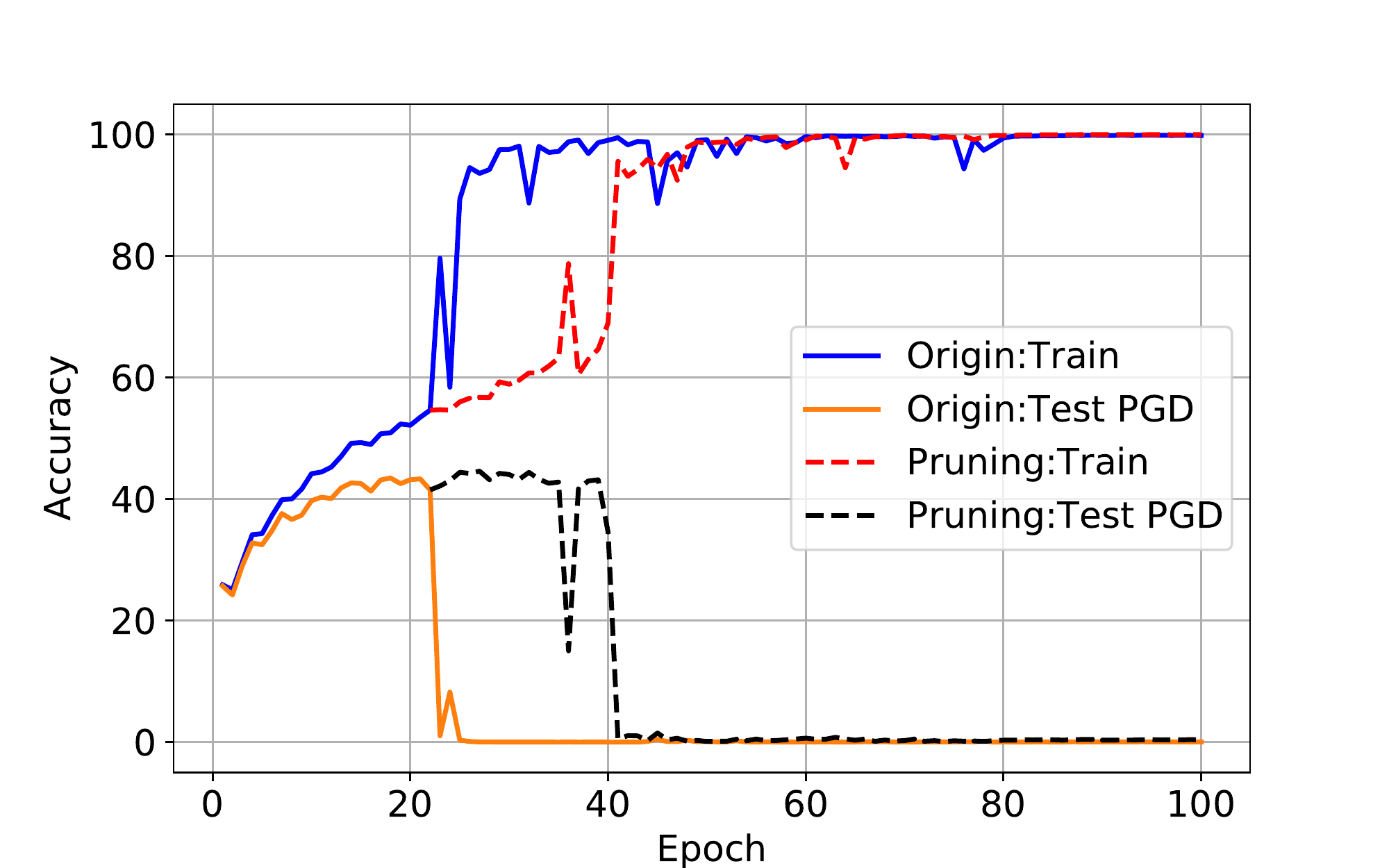}
\caption{Training curves of original training and the network with CO after pruning high-variance channels on Cifar10 using ResNet18 with $\epsilon=8/255$. After pruning, the network can regain PGD robustness for some epochs. }
\label{fig:retrain}
\end{figure}

To further observe the effect of high-variance channels on training, we continue the training after setting the parameters associated with the pruned channel with the highest variance to zero. Fig.\ref{fig:retrain} shows that pruned high-variance channels can help the network recover robustness quickly after a short training and delay the occurrence of CO.

\section{Further Discussion}
\label{initialization}

\textbf{Is adversarial initialization sufficient to prevent FAT from falling into self-fitting?} 
Some works  \cite{sriramanan2020guided, jia2022prior,chen2022efficient} claim that better initialization can help to generate more adversarial examples and then mitigate the occurrence of CO. But from the perspective of self-fitting, simply improving the initialization without applying regularization to the training process cannot prevent networks from recognizing the self-information. When enough self-information is added on inputs, i.e., $\epsilon$ is large enough, the network could still find the shortcut solution and fall into self-fitting. 

\begin{figure}[ht]
\centering
\includegraphics[width=0.42\textwidth]{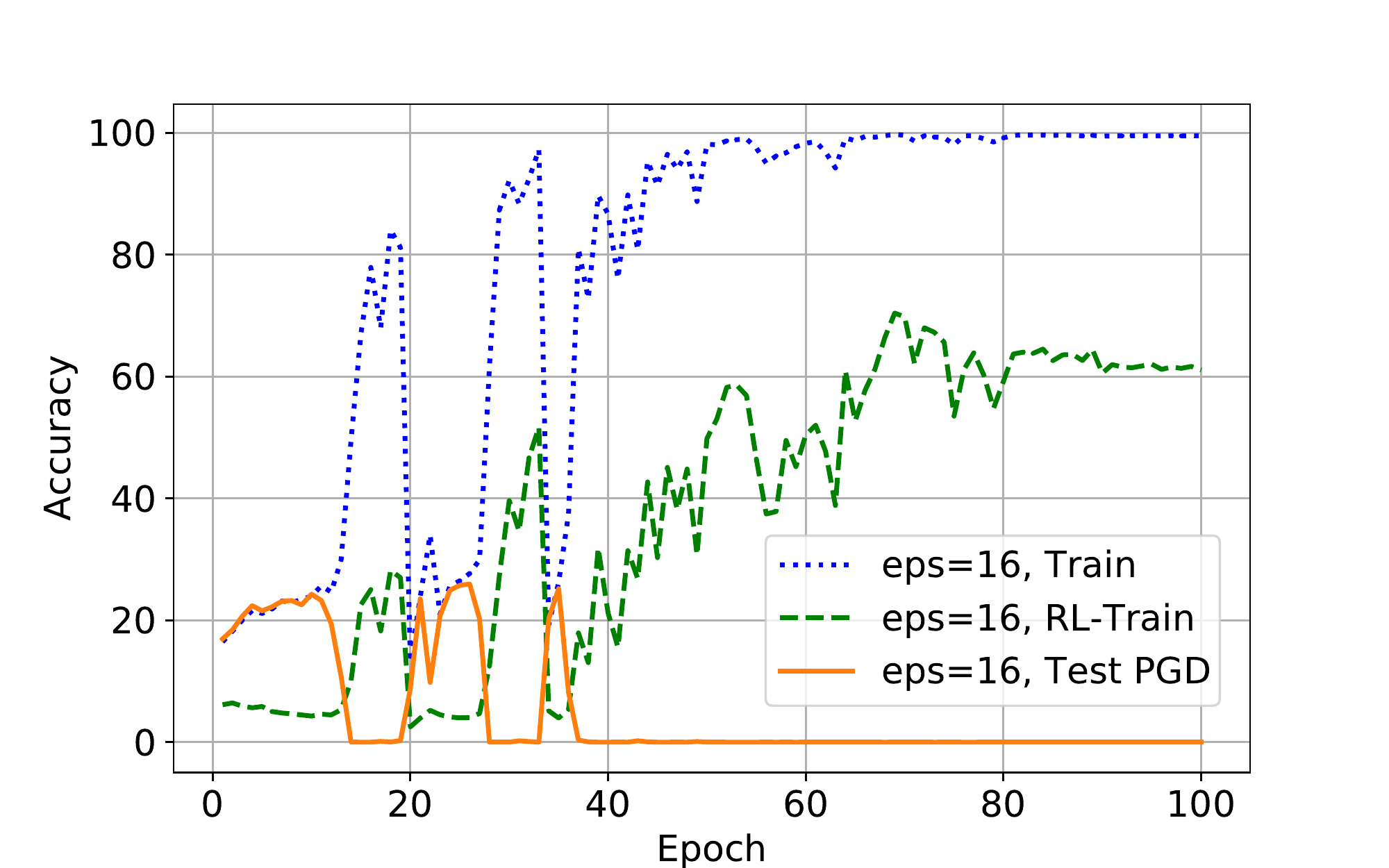}
\caption{FGSM-AT with PGD-7 perturbation initialization. We observe catastrophic overfitting still happens}
\label{fig:pgd-init}
\end{figure}

To explore whether adversarial initialization can mitigate CO, a ResNet18 is trained on Cifar10 with $\epsilon=16$ following the experiment settings in \cref{fig:base-training}. The only difference is that a PGD attack with 7 steps is used here to find an adversarial initialization before adding the single-step perturbation. 
\cref{fig:pgd-init} shows that even with a sufficiently adversarial initialization, CO stills happens when $\epsilon$ is large enough. Compared to training with the random initialization, training with PGD-7 as initialization is less stable and can regain robustness sometimes. However, after long enough training, CO still happens with such an initialization.

\textbf{Regularization can help mitigate CO.}
Some works add proper regularization terms in  loss function to guide the training, such as  GradAlign \cite{andriushchenko2020understanding}  and NuAT \cite{sriramanan2021towards}.
From the perspective of self-fitting, these regularization terms prevent the network from directly using self-information for classification by matching certain properties of original examples with those of adversarial examples. 
To some extent, these regularization terms interfere with the learning of self-information while helping identify clean samples or randomly perturbed samples, thereby avoiding  CO.

For example, \cite{andriushchenko2020understanding} found that after CO a dramatic decrease in local linearity happens. After CO the network turns to recognize self-information instead of data information. As a consequence, even though the original data point $x$ and the data point after the random perturbation $x + \delta_0$ are close on the original data manifold, they are far from each other on the self-information manifold. The gradients at these two points are nearly orthogonal to each other. Thus, increasing the 
cosine similarity of $x$ and $x + \delta_0$  can exclude the influence of self-information and the mitigate CO. 
 We also observe that aligning the gradient of $x$ and $x_{\text{FGSM}}$ cannot prevent CO, since in this regularization term self-information in $x_{\text{FGSM}}$ can be captured by the network.

\begin{figure}[ht]
\centering
\includegraphics[width=0.35\textwidth]{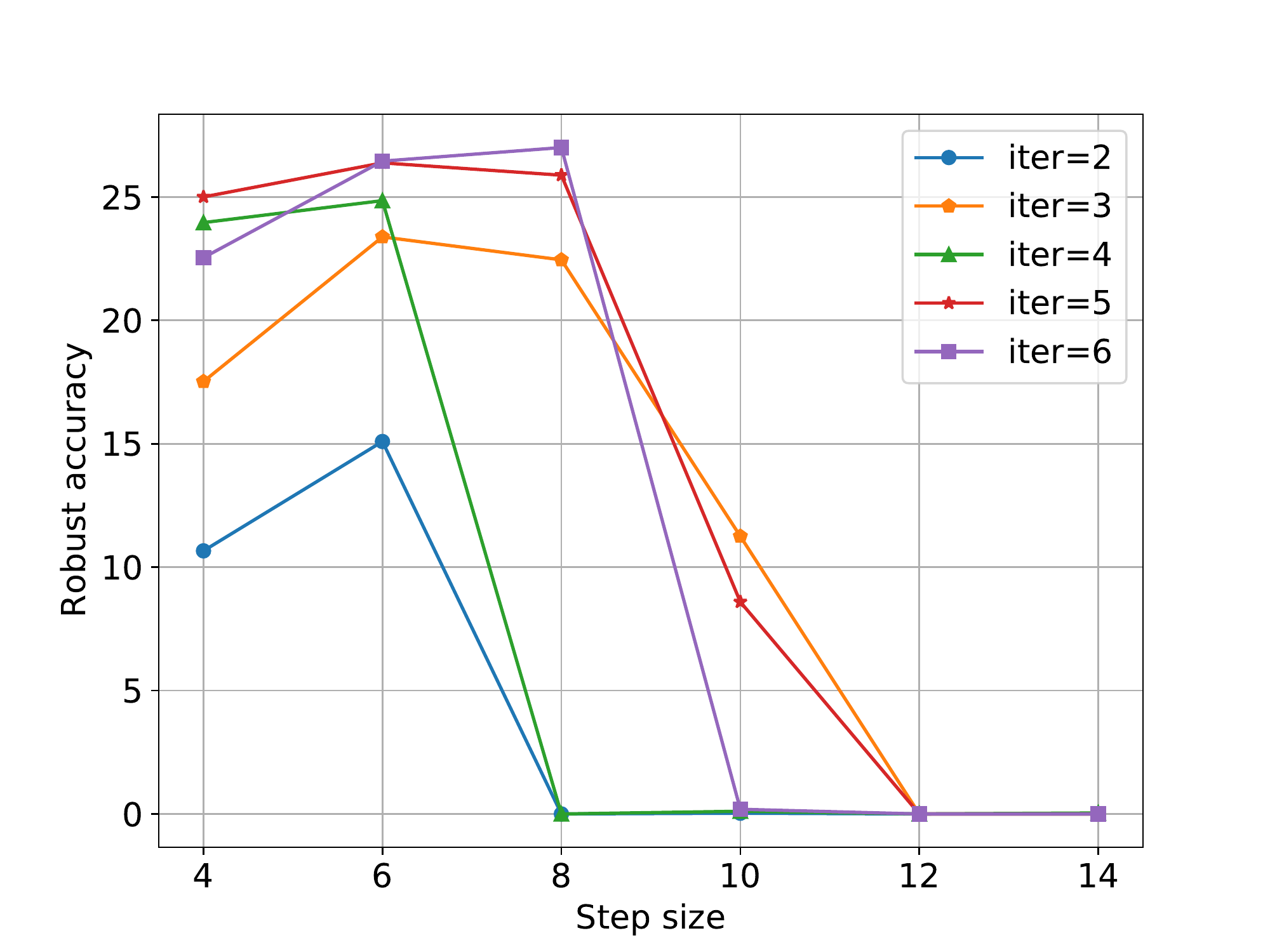}
\caption{Robust accuracy of multi-step AT with different iterations and step sizes. $\epsilon=16/255$ and the robust accuracy is the minimum value of the last half training, tested with a PGD-10 attack with random start.}
\label{fig:pgd}
\end{figure}

\textbf{CO also happens in multi-step adversarial training.}
The results above focus on self-fitting in FAT, but self-fitting should also happen in multi-step adversarial training.
To explore whether CO would happen in multi-step AT, we train ResNet18 on Cifar10 with different iterations and step sizes using a PGD attack.
The results in \cref{fig:pgd} suggest that in some settings, the test robust accuracy would drop to nearly 0, which means CO still happens in multi-step AT as expected. 
Firstly, for large step size $\alpha$($\alpha \geq 12/255$), increasing iterations cannot prevent networks from CO. 
Secondly, by fixing the iteration number, the robust accuracy rate rises and then falls as  $\alpha$ increases. When $\alpha$ is small, increasing it can generate more adversarial examples. When $\alpha$ is large, increasing it leads to the decrease of robust accuracy, with first robust overfitting and finally catastrophic overfitting.

\section{Conclusion}
In this paper, we propose a novel perspective, i.e.,\emph{ self-fitting},  to interpret catastrophic overfitting in fast adversarial training. Self-fitting reveals a possible shortcut solution in adversarial training that an AT-trained DNN can embed self-information into adversarial perturbation for classification while ignoring data-information. 
Moreover, we also find a \emph{"channel differentiation"} phenomenon that different channels of the network’s first layer recognize different types of examples, which gives further evidence that part of the network is dedicated to recognizing self-information. 
Based on self-fitting, we can explain the existing method to mitigate CO and extend CO to multi-step AT. We believe that the interaction between the model and adversarial perturbation is an important reason for CO. 
Our findings provide a new perspective on CO prevention that how to preserve the adversarial properties of AEs while reducing the influence of self-information.


\pagebreak
{\small
\bibliographystyle{ieee_fullname}
\bibliography{egbib}
}

\appendix   

\appendix   
\setcounter{table}{0}   
\setcounter{figure}{0}
\renewcommand{\thetable}{A\arabic{table}}
\renewcommand{\thefigure}{A\arabic{figure}}

\twocolumn[
\begin{@twocolumnfalse}
\section*{\centering{Supplementary Material for \\ \emph{Investigating Catastrophic Overfitting in Fast Adversarial Training: A Self-fitting Perspective}} \\~}
\end{@twocolumnfalse}
]

\section{Experiment details.} \label{appendix:experiment-details}
\textbf{FAT settings.}
We train ResNet18 on Cifar10 with the FGSM-AT method  \cite{wong2020fast} for 100 epochs in Pytorch \cite{NEURIPS2019_9015}. We set $\epsilon = 8/255$ and $\epsilon=16/255$ and use a SGD  \cite{qian1999momentum} optimizer with 0.1 learning rate. The learning rate decays with a factor of 0.1 at the 80th and 90th epochs. To better study CO, we use zero initialization to generate adversarial samples, and weight decay is set to 0 to reproduce CO stably. The batch size is 128. Images are padded with 4 pixels and randomly cropped and flipped horizontally.

\begin{figure}[ht]
\centering
\includegraphics[width=0.45\textwidth]{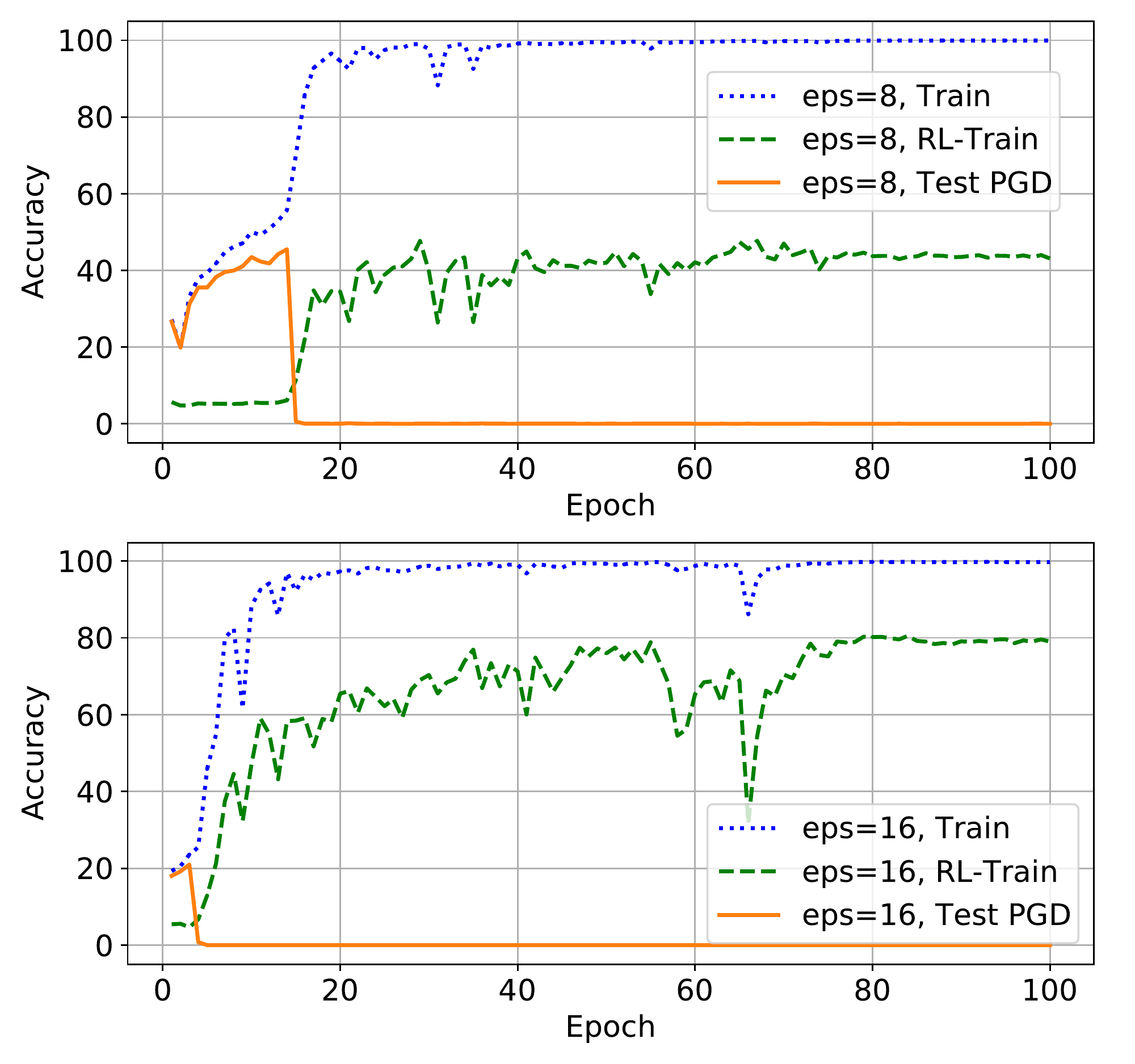}
\caption{FGSM-AT training with different $\epsilon$ on Cifar10 using WideResNet28-10. Catastrophic overfitting happens at 15th epoch for $\epsilon = 8/255$ and 4th epoch for $\epsilon = 16/255$. }
\label{fig:base-training-wideresnet}
\end{figure}

\begin{figure*}[!t]
\centering
\includegraphics[width=1\textwidth]{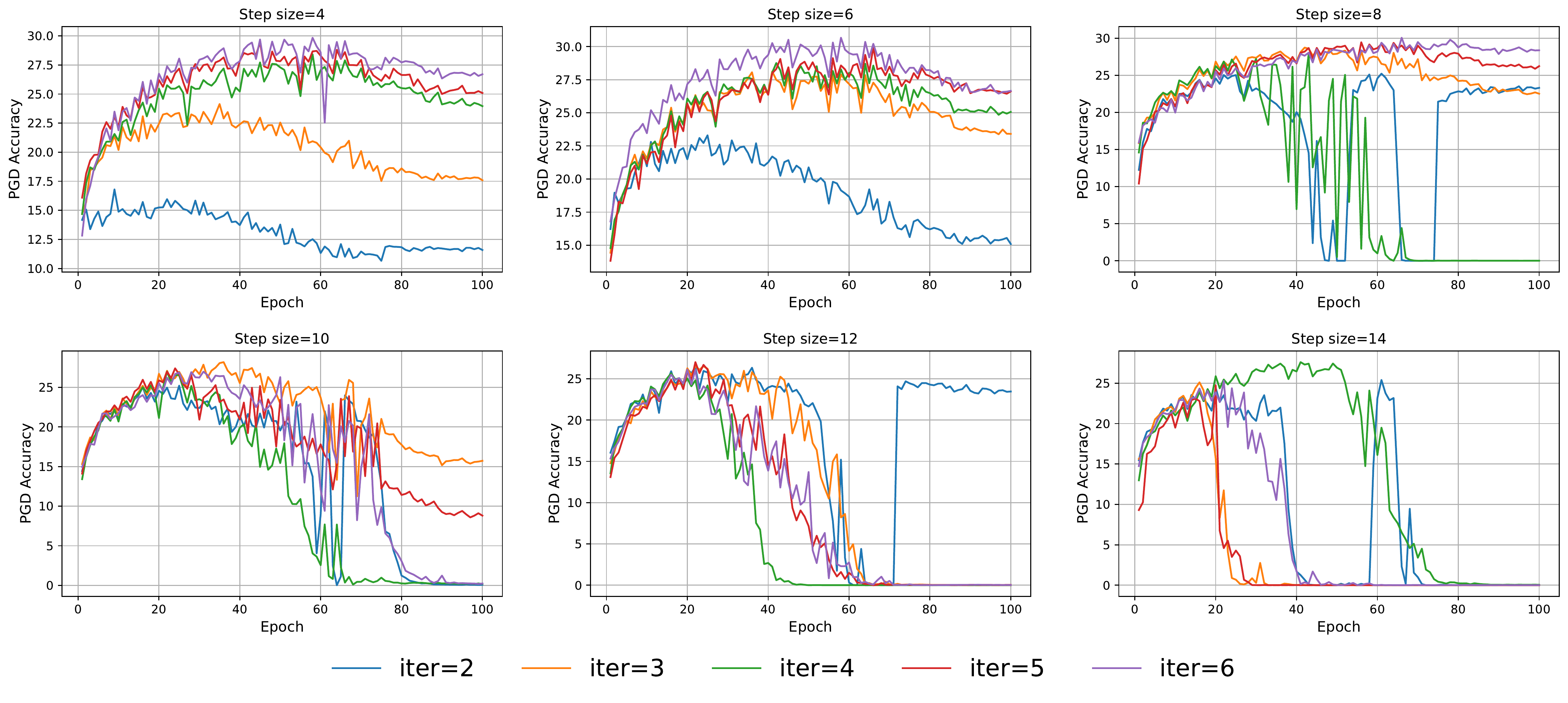}
\caption{Training curves of multi-step AT with different iterations and step sizes. PGD accuracy is calculated using a PGD20 attack with 3 random starts.}
\label{fig:pgd_detail}
\end{figure*}

\textbf{PGD-AT details in further discussion.}  There is only a little difference between the settings of PGD-AT and FAT. 
PGD-AT uses a smaller step size and more iterations with $\epsilon=16/255$. 
The learning rate decays at the 75th and 90th epochs.
The robust accuracy during training of different settings is shown in \cref{fig:pgd_detail}.

\section{Experiments on WideResnet28-10.}
This section reports  experiments on WideResNet28-10 about self-fitting. Compared to ResNet18, WideResNet28-10 has more parameters and therefore has a stronger learning capability.

\textbf{Training curve.}
\cref{fig:base-training-wideresnet} shows the training curve of WideResNet28-10 on Cifar10 with FGSM-AT method. The training setting also follows \cref{appendix:experiment-details}. Catastrophic overfitting happens earlier than ResNet18.
After CO, the random-label FGSM accuracy also increases quickly with training accuracy, suggesting that self-information dominates the classification.

\begin{figure}[ht]
\centering
\includegraphics[width=0.48\textwidth]{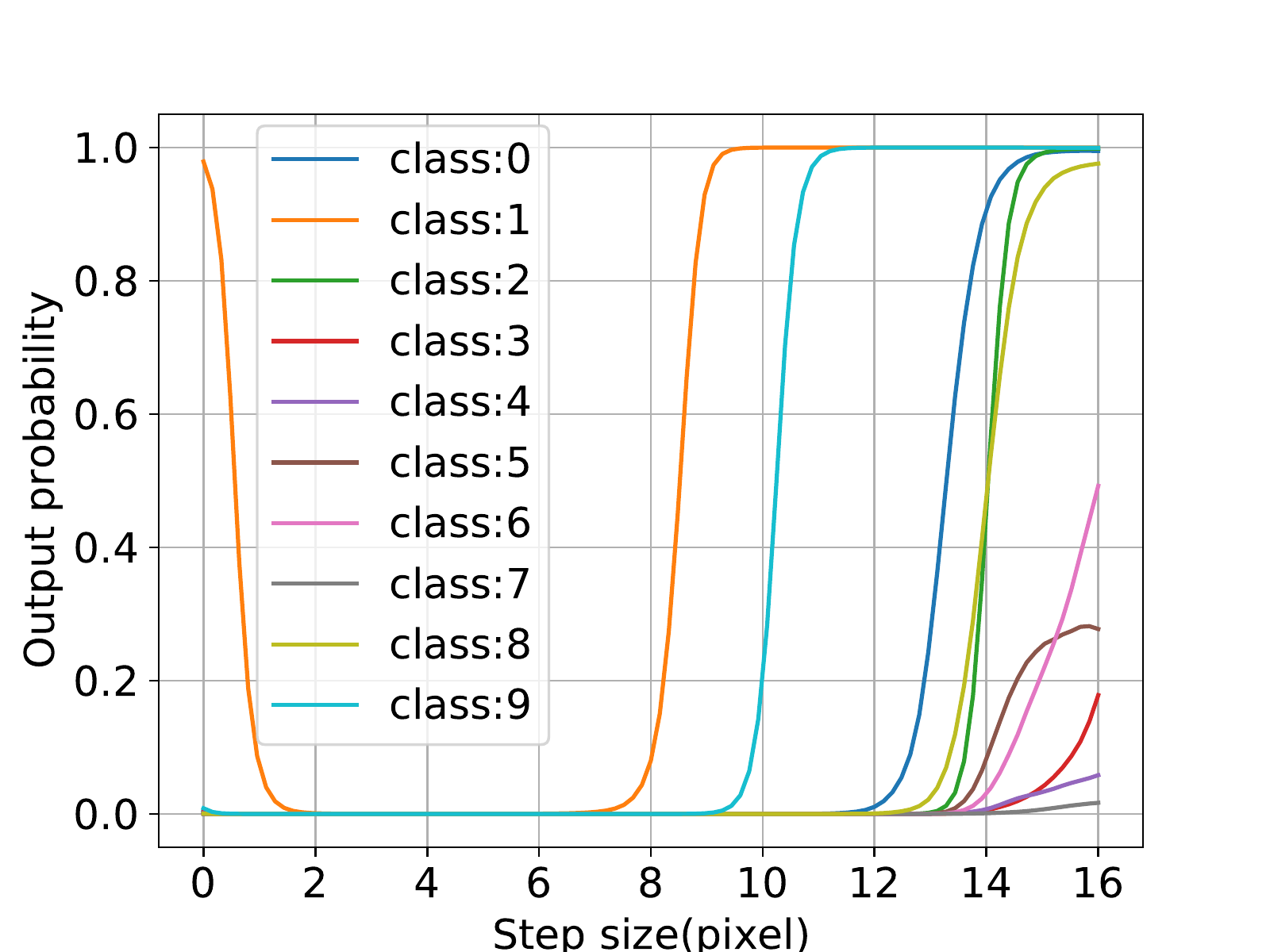}
\caption{Visualization of the probability of the network in the corresponding classes when the step size of FGSM perturbation gradually increases. The original class is class 1. The network is trained with $\epsilon = 16/255$. }
\label{fig:change-alpha-wideresnet}
\end{figure}

\textbf{Probability changes with attack step size's increase.}
 \cref{fig:change-alpha-wideresnet} visualizes that when the step size of FGSM perturbation gradually increases, how output probability of the network in the corresponding classes. The model is trained on Cifar10 using WideResNet28-10.
 When the step size increases, the probability firstly decreases, meaning that the perturbation can fool the network, then increases, meaning that the network can recognize the self-information in the perturbation when the step size is large enough.
 
 \begin{table}[ht]
\caption{Accuracy drop of different networks trained on Cifar10 using WideResNet28-10. Compared  to the network without CO, the network with CO has a large drop in FGSM accuracy while little change in clean accuracy.}
\label{tabel:accuracy-change-wideresnet}
\vskip 0.15in
\begin{center}
\begin{small}
\begin{sc}
\begin{tabular}{crrr}
\toprule
                      & Clean & FGSM & PGD  \\ 
\midrule
w/o CO, not pruned & 76.4\%   & 51.2\% & 45.4\%   \\
w/o CO, pruned     & -19.9\%    & -19.5\%  & -16.2\%  \\ 
\midrule
 with CO, not pruned    & 79.2\%    & 99.5\%   & 0.0\%  \\
 with CO, pruned        & -10.4\%    & -92.4\%   & 0.1\% \\
\bottomrule
\end{tabular}
\end{sc}
\end{small}
\end{center}
\vskip -0.1in
\end{table}

\begin{figure}[ht]
\centering
\includegraphics[width=0.5\textwidth]{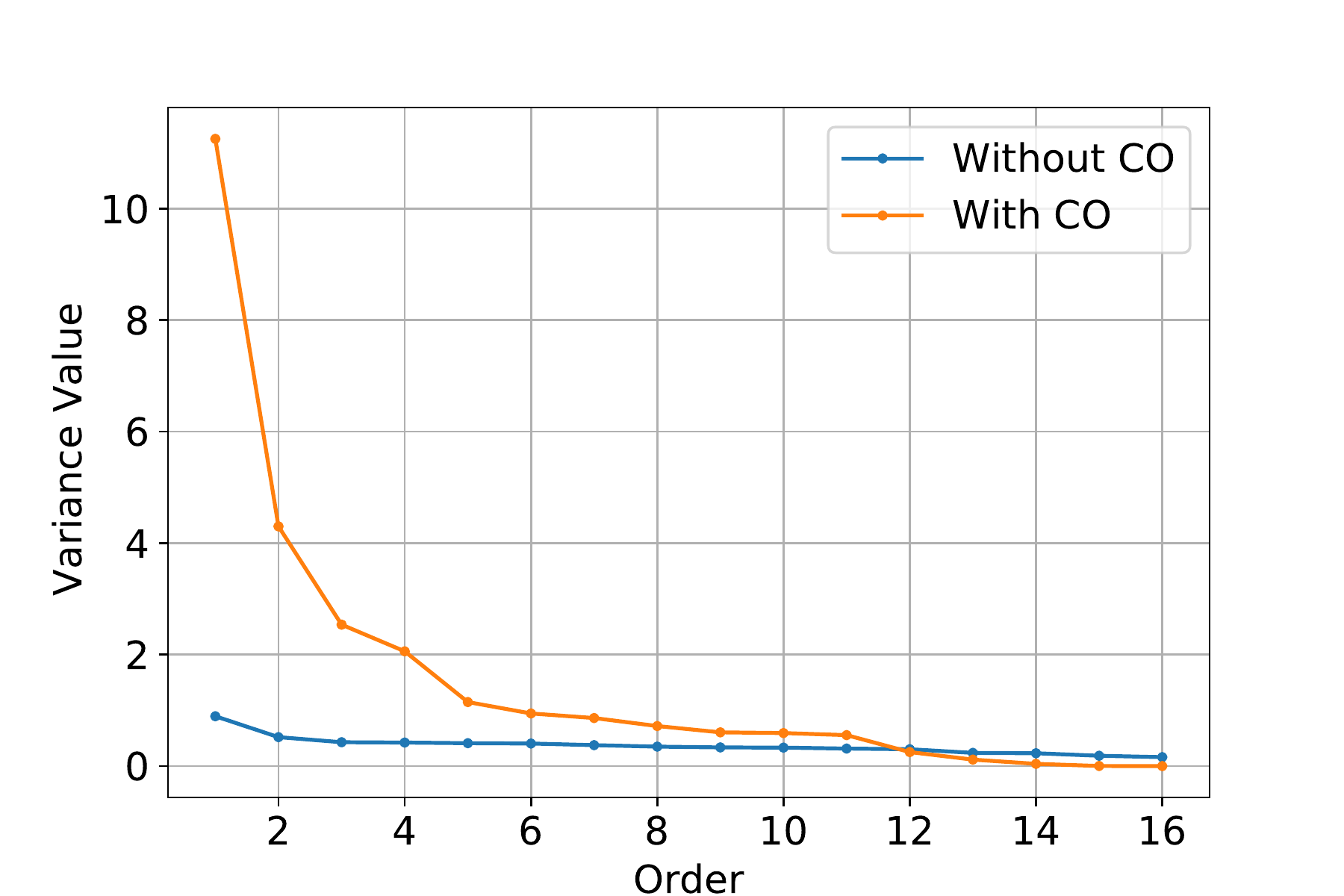}
\caption{The variance values in descending order of networks with and without the CO on WideResNet28-10. The network with CO has a larger maximum variance value and more zero variance channels. }
\label{fig:channel-variance-wideresnet}
\end{figure}

\textbf{Channel variance in descending order.}
\cref{fig:channel-variance-wideresnet} shows the variance values in descending order of networks with and without the CO on WideResNet28-10. The features after the first layer of WideResNet28-10 have only 16 channels.
After CO, some channels become dominant to recognize self-information, thus having a larger variance. While some channels for data-information become unimportant and ``dead''.

\textbf{Accuracy of pruned network.}
\cref{tabel:accuracy-change-wideresnet} shows the accuracy change of different setting 
after pruning, which is for WideResNet28-10 trained on Cifar10. Only one channel of the first layer with the highest variance is pruned. The network without CO has a similar drop in all accuracy after pruning. In contrast, after pruning, the network with CO has a drop of 92.4\% in FGSM accuracy, while the clean accuracy only decreases by 10.4\%.


\end{document}